\documentclass[letterpaper, 10 pt, conference]{ieeeconf}

\IEEEoverridecommandlockouts
\usepackage[T1]{fontenc}
\usepackage{graphicx}
\usepackage{amsmath}
\usepackage{amssymb}
\usepackage{booktabs}       
\usepackage{multirow}
\usepackage{url}
\usepackage{hyperref}
\usepackage[table]{xcolor}  
\usepackage[font=footnotesize]{caption}  
\usepackage{subcaption}     
\usepackage{xspace}
\usepackage{ifthen}
\usepackage{siunitx}

\DeclareMathOperator{\arctantwo}{atan2}
\newcommand{\norm}[1]{\lVert #1 \rVert}

\newcommand{\Cpp}{C\raise.08ex\hbox{\tt ++}}

\newcommand{\SonoRank}{\textsf{SonoRank}}
\newcommand{\MFT}{\textsf{MFT}}
\newcommand{\MFTRest}{\textsf{MFT{+}Rest}}

\newcommand{\BCE}{\text{BCE}}

\newcommand{\ignore}[1]{}

\newboolean{HIDENOTES}
\setboolean{HIDENOTES}{false}

\ifthenelse{\boolean{HIDENOTES}}{
    \newcommand{\OS}[1]{{}}
    \newcommand{\DZ}[1]{{}}
    \newcommand{\AB}[1]{{}}
    \newcommand{\AW}[1]{{}}
    \newcommand{\CONT}[1]{{}}
    \newcommand{\OPUS}[1]{{}}
    
}{
    \newcommand{\OS}[1]{\textcolor{red}{#1}}
    \newcommand{\DZ}[1]{\textcolor{blue}{#1}}
    \newcommand{\AB}[1]{{\textcolor{green}{#1}}}
    \newcommand{\AW}[1]{{\textcolor{yellow}{#1}}}

    \newcommand{\OPUS}[1]{{\textcolor[RGB]{181,101,29}{#1}}}
    
}

\let\labelindent\relax        
\usepackage{enumitem}
\setlist{nosep, leftmargin=*}

\title{\LARGE \bf
SonoRank: Towards Calibration-Free Real-Time Finger Flexion Detection from Forearm Ultrasound Sequences
}

\author{Dean Zadok$^{1}$, Alon Wolf$^{2}$, Alex M. Bronstein$^{1}$, Oren Salzman$^{1}$
\thanks{$^{1}$Department of Computer Science, Technion, Haifa, Israel
        {\tt\small \{deanzadok,bron,osalzman\}@cs.technion.ac.il}}%
\thanks{$^{2}$Department of Mechanical Engineering, Technion, Haifa, Israel
        {\tt\small alonw@me.technion.ac.il}}%
}

\begin{document}

\maketitle
\thispagestyle{empty}
\pagestyle{empty}

%

\begin{abstract}

Powered prosthetic hands are frequently abandoned, largely due to the limited functionality of current devices that rely on surface electromyography (sEMG).
Sonomyography (ultrasound) has emerged as a promising alternative, owing to its ability to observe muscle activity in real time and control a greater number of degrees of freedom.
Yet, existing ultrasound-based methods require per-user fine-tuning, limiting their commercialization.
We propose \SonoRank{}, an important step towards calibration-free finger flexion detection from forearm ultrasound video.
\SonoRank{} first learns to rank pairs of ultrasound sequences by their relative motion magnitude for each of the five fingers.
The learned representations are then fine-tuned to classify whether each finger is actively flexing, using a rest reference that is captured at the beginning of the operation.
Under 12-fold leave-one-subject-out cross-validation on a dataset of twelve subjects with synchronized kinematics, \SonoRank{} achieves a $28$\% improvement in F1 score over direct classification baselines that skip the ranking stage.
These results establish pairwise ranking as an effective pretraining signal for subject-independent control, bringing ultrasound-based prosthetics closer to practical, calibration-free deployment.

\end{abstract}

\section{INTRODUCTION}
\label{sec:intro}

Six decades of prosthetics research have transformed artificial hands from simple grippers into mechatronic systems.
Modern designs offer dozens of independent degrees of freedom, approaching human-level dexterity~\cite{zadok2026digiarm}.
Yet 44\% of upper-limb amputees reject their prostheses~\cite{salminger2022acceptance}, suggesting a gap in the control interface.
Since the mid-twentieth century, surface electromyography (sEMG) has remained the dominant non-invasive sensing modality for prosthetic hands~\cite{herberts1969myoelectric}, alongside plug-and-play interfaces~\cite{zadok2025ankleband}.
However, sEMG captures only noisy electrical activity from superficial muscle fibers, making individual finger discrimination difficult~\cite{DBLP:conf/hsi/HuangL16}.
Montagnani et al.\ showed that it is precisely finger dexterity that current prostheses lack~\cite{montagnani2015finger}, underscoring that the bottleneck lies mostly in sensing.

\begin{figure}[t]
\centering
\includegraphics[width=0.875\columnwidth]{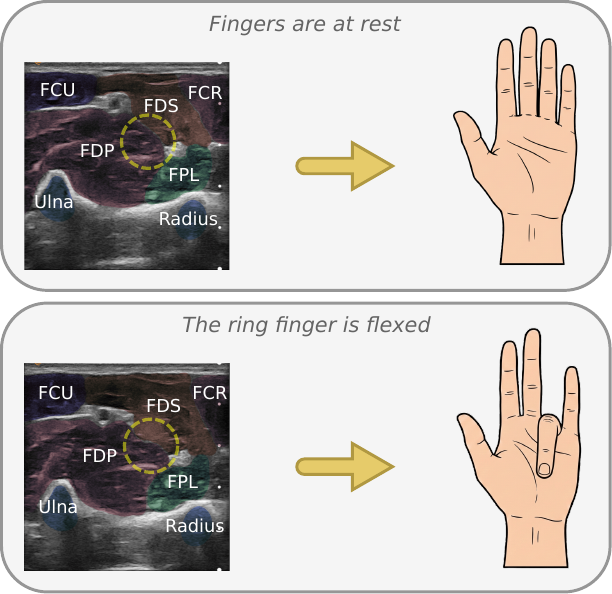}
\vspace{-0.025cm}
\caption{Ultrasound can see what EMG cannot: B-mode cross-sections of the forearm (top of each image is the skin surface) at rest (top) and during ring-finger flexion (bottom), with the corresponding hand pose on the right.
Muscle regions and bones were manually annotated for illustration purposes: the flexor digitorum superficialis (FDS) and profundus (FDP), the two extrinsic muscles that actuate the four fingers, are highlighted alongside the Ulna and Radius (blue).
The dashed circle highlights the boundary between the FDS and FDP. During flexion, this region visibly deforms as the muscles contract and shift relative to each other.
This morphological change is the signal that our method is designed to detect.}
\vspace{-0.4cm}
\label{fig:anatomy}
\end{figure}

%

Ultrasound imaging of the forearm (i.e., sonomyography) offers a richer window into motor intent~\cite{10816243}.
B-mode (brightness-mode) images, the grayscale cross-sections produced by conventional ultrasound, capture the spatial structure of deep muscles in real time, enabling the discrimination of finger movements from morphological deformations that sEMG cannot resolve~\cite{DBLP:conf/iros/CastelliniP11,DBLP:journals/tbe/SgambatoHBIJFTF24}.
Yet existing sonomyography methods frame finger motion prediction as either discrete gesture classification~\cite{DBLP:conf/chi/McIntoshMFP17,lee2025fingermotion} or direct joint-angle regression~\cite{DBLP:journals/tbe/SgambatoHBIJFTF24,DBLP:journals/corr/abs-2211-15871}.
Classification restricts output to a fixed gesture vocabulary, while regression demands subject-specific calibration of continuous kinematics labels~\cite{cerveri2007finger}.
Moreover, most approaches treat each finger independently, ignoring the biomechanical coupling among forearm muscles that actuate multiple fingers simultaneously~\cite{montagnani2015finger}.
Cross-subject generalization has so far been limited to A-mode armbands for wrist and hand tracking~\cite{sgambato2025vr} or to systems that fuse ultrasound with surface electromyography~\cite{bastola2025hybrid}.
The key question this work aims to answer is: \emph{can we detect individual finger flexions from forearm ultrasound and instruct a robotic hand to replicate them, without calibration?}\footnote{We do not showcase results on deformed muscles. We test on healthy subjects with fully functioning hands to provide an accurate benchmark for this task. Our code is available at \textcolor{blue}{\href{https://github.com/deanzadok/sonorank}{github.com/deanzadok/sonorank}}.}

%

\begin{figure*}[t]
    \centering
    \includegraphics[width=0.975\textwidth]{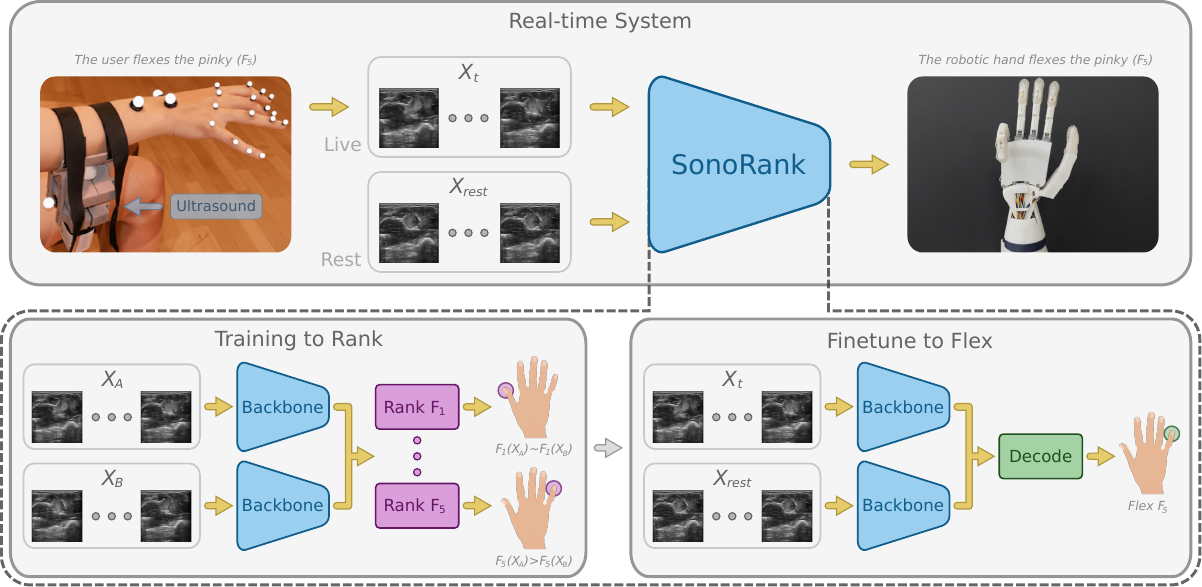}
    \caption{Overview of the \SonoRank{} framework.
    (Top)~Real-time operation: a forearm ultrasound probe captures a live sequence $X_t$ while the user flexes a finger.
    Together with a pre-recorded rest reference $X_{\text{rest}}$, both sequences are fed into \SonoRank{}, which outputs per-finger flexion decisions that drive a robotic hand.
    (Bottom left)~Stage~1, training to rank: two ultrasound sequences $X_A$ and $X_B$ are encoded through a shared backbone, and five parallel ranking heads (one per finger, $F_1, \ldots, F_5$) learn to predict which sequence exhibits greater flexion for each finger.
    (Bottom right)~Stage~2, fine-tuning to flex: the ranking heads are replaced by a classification head (Decode) that receives the concatenated embeddings of $X_t$ and $X_{\text{rest}}$ and predicts which fingers are actively flexing.
    }
    \vspace{-0.2cm}
    \label{fig:overview}
\end{figure*}

Towards this goal, we propose \SonoRank{}, which decomposes finger flexion detection into two training stages.
First, a ranking model learns to compare pairs of ultrasound sequences by their relative motion magnitude for all five fingers independently.
Because ranking compares relative motion rather than absolute measurements, the learned representations generalize across forearm morphologies without calibration.
A second stage then fine-tunes the model to decide, for each finger, whether it is actively flexing.
To the best of our knowledge, this is the first work to use pairwise ranking as a pretraining signal for subject-independent finger flexion detection from forearm ultrasound.
Our contributions are:
(i) A framework for cross-subject finger flexion detection from forearm ultrasound that requires no user-specific data or fine-tuning at deployment,
(ii) A pairwise ranking formulation for sonomyography that learns subject-independent motion representations by comparing flexion magnitudes, avoiding calibration across different forearm morphologies, and
(iii) Lessons learned from offline and online studies on experimenting with a calibration-free robotic system for the first time.
Under 12-fold leave-one-subject-out cross-validation, \SonoRank{} improves over direct classification baselines and outperforms all external methods by a wide margin, while running in real time on affordable GPUs.

%



\section{RELATED WORK}
\label{sec:related}



Sonomyography has progressed from initial feasibility to wearable real-time systems over the past fifteen years~\cite{10816243}.
Castellini and Passig~\cite{DBLP:conf/iros/CastelliniP11} first showed that forearm ultrasound features are linearly related to finger positions, and subsequent work demonstrated real-time gesture classification~\cite{DBLP:conf/chi/McIntoshMFP17,DBLP:conf/biorob/CastelliniHSGN14}, confirming that US matches or exceeds sEMG for finger discrimination~\cite{DBLP:conf/hsi/HuangL16}.
Since then, sonomyographic control has been validated clinically in individuals with upper-limb loss~\cite{DBLP:journals/corr/abs-1808-06543,nazari2025highly}, extended to nine individual finger motions~\cite{lee2025fingermotion}, and demonstrated for cross-participant hand tracking~\cite{sgambato2025vr,DBLP:journals/tbe/SgambatoHBIJFTF24}.
Wearable probes~\cite{yan2019lightweight,DBLP:journals/tsmc/YangCHYL21} and flexible transducers~\cite{peng2023transformer} have moved the technology toward deployment, while parallel efforts address sensor-reattachment robustness~\cite{DBLP:journals/titb/YangZZHL19}, cross-subject transfer~\cite{transferus2024}, ultra-low-power inference~\cite{vostrikov2024unsupervised}, and clinical prosthetic validation~\cite{smgreview2024}.

Single-frame approaches, including early sonomyographic prosthetic control systems~\cite{DBLP:conf/embc/BimbrawFWH20}, discard the temporal dynamics of muscle contraction visible in ultrasound video.
In the broader biosignal literature, hybrid CNN-Transformer architectures have become a dominant paradigm for temporal decoding~\cite{anwar2025transformers}.
Zadok et al.~\cite{DBLP:conf/icra/ZadokSWB23,zadok2025inferring} introduced a temporal architecture with kinematic representations for predicting fine finger motions from ultrasound, while Jiang et al.~\cite{peng2023transformer} combined a flexible US transducer with a Sonomyography Transformer for simultaneous gesture and force recognition.
In the EMG domain, Zhang et al.~\cite{zhang2020simultaneous} estimated multiple joint kinematics simultaneously, and Weng et al.~\cite{fingermotion2025amputees} demonstrated real-time fine finger motion decoding in amputees.
Despite these advances, most methods frame finger motion prediction as classification or regression, without exploiting relative temporal comparisons.

Pairwise ranking formulations have found applications beyond their origins in information retrieval.
Misra et al.~\cite{misra2016shuffle} showed that temporal order verification in video provides a self-supervised signal for visual representation learning.
In medical imaging, Li et al.~\cite{li2020siamese} used Siamese networks with pairwise comparison against references to produce continuous disease severity scores, and Dong et al.~\cite{dong2021deepatrophy} showed that ranking longitudinal MRI scan pairs by temporal order captures progressive brain atrophy.
Pairwise ranking has also been applied to skill assessment and temporal change detection in video~\cite{doughty2018skill}, and to time-series forecasting, where temporal relational ranking outperforms pointwise regression~\cite{feng2019temporal}.
Our work follows a similar motivation: we use pairwise temporal ranking as a supervised learning signal, comparing motion sequences against rest references to generalize across unseen forearm morphologies.
To jointly predict all five fingers from a shared backbone, we adopt multi-task learning with a shared representation and per-finger loss balancing~\cite{kendall2018multi}.
\section{METHOD}
\label{sec:method}


We begin with the anatomy linking forearm ultrasound to finger motion.
The forearm contains several extrinsic flexor muscles whose contractions are visible in B-mode ultrasound and directly control finger motion~\cite{lieber1992forearm}.
The flexor digitorum superficialis (FDS) and flexor digitorum profundus (FDP) actuate the index, middle, ring, and pinky fingers through tendons that cross the proximal interphalangeal (PIP) and distal interphalangeal (DIP) joints, respectively, while the flexor pollicis longus (FPL) controls the thumb.
Because the FDS and FDP share tendons across the four fingers, flexing one finger produces morphological changes that overlap with those of neighboring fingers, making per-finger discrimination challenging (Fig.~\ref{fig:anatomy}).
The FPL is one of the smallest extrinsic flexors, and the primary thumb muscles (the thenar group) lie in the palm, outside the ultrasound field of view, which complicates thumb prediction.
These anatomical constraints motivate a shared multi-finger and temporal representation that can learn the coupled muscle dynamics.

\subsection{Problem Definition}
\label{sec:formulation}

Let $x_t \in \mathbb{R}^{H \times W}$ denote a B-mode ultrasound image of the forearm acquired at time~$t$ (see Fig.~\ref{fig:overview}).
For each finger $f \in \mathcal{F} = \{F_1, \ldots, F_5\}$ (thumb, index, middle, ring, pinky), let $\theta^{(f)}_{\text{MP},t}$ and $\theta^{(f)}_{\text{PIP},t}$ denote its metacarpophalangeal and proximal interphalangeal joint angles at time~$t$, and define the total flexion angle as $\theta^{(f)}_t = \theta^{(f)}_{\text{MP},t} + \theta^{(f)}_{\text{PIP},t}$.
A temporal sequence $X = \{x_1, \ldots, x_T\}$ consists of $T$ consecutive frames from a recording trial.
We define the motion magnitude of finger~$f$ over the sequence as the absolute change in flexion angle:
\begin{equation}
  \delta^{(f)}(X) = \bigl|\theta^{(f)}_{T} - \theta^{(f)}_{1}\bigr|.
  \label{eq:delta}
\end{equation}
Our goal is to predict whether each finger is actively flexing within a window of frames.
We decompose this problem into two stages: a pairwise ranking model that learns motion dominance for each finger from pairs of sequences (Sec.~\ref{sec:ranking}), and a classifier that predicts finger flexion from the learned representations (Sec.~\ref{sec:flex_classification}).
We refer to the combined two-stage system as \SonoRank{}.

\subsection{Pairwise Temporal Ranking}
\label{sec:ranking}

Rather than predicting flexion labels directly, we first train a model to compare temporal sequences by their relative motion magnitude.
Given two sequences $X_A, X_B$ from the same trial, the ranking label for finger~$f$ is
\begin{equation}
  y^{(f)} = \mathbf{1}\!\bigl[\delta^{(f)}(X_A) > \delta^{(f)}(X_B)\bigr].
  \label{eq:label}
\end{equation}
Additionally, a pair is \emph{informative} for finger~$f$ only when the motion difference exceeds a minimum threshold~$\tau_{\min}$:
\begin{equation}
  m^{(f)} = \mathbf{1}\!\bigl[\bigl|\delta^{(f)}(X_A)
    - \delta^{(f)}(X_B)\bigr| \geq \tau_{\min}\bigr],
  \label{eq:active}
\end{equation}
with $\tau_{\min} = \SI{3}{\degree}$ in all experiments.
The ranking model (Fig.~\ref{fig:overview}, bottom left) consists of a shared encoder that maps each ultrasound frame to a feature vector, a temporal transformer module that aggregates frame-level features into a sequence-level embedding, and five parallel comparison heads (one per finger $f$) that each produce the ranking probability
\begin{equation}
  \hat{p}^{(f)} = P\bigl(\delta^{(f)}\!(X_A) > \delta^{(f)}\!(X_B) \mid X_A, X_B\bigr).
  \label{eq:ranking_prob}
\end{equation}
%
%
The shared encoder and transformer capture dynamics common to all fingers, while each comparison head specializes in the motion pattern of each finger.
The training objective combines binary cross-entropy ($\ell_{\text{BCE}}$) on informative pairs with an uncertainty penalty on uninformative ones:
\begin{equation}
  \mathcal{L}_{\text{rank}} = \frac{1}{|\mathcal{F}|}\sum_{f \in \mathcal{F}} \Bigl[
    m^{(f)} \ell_{\text{BCE}}^{(f)}
    + \lambda\, (1{-}m^{(f)})\, \bigl(\text{logit}^{(f)}\bigr)^{\!2}
  \Bigr].
  \label{eq:loss}
\end{equation}
The first term trains the model to rank correctly when sufficient motion difference exists.
The second term is an \emph{uncertainty penalty} applied when the finger~$f$ is inactive ($m^{(f)}{=}0$). The squared logit pushes predictions toward $\hat{p}{=}0.5$, preventing the model from learning unwanted rankings on pairs where motion ordering is ambiguous.
This ranking formulation avoids subject-specific calibration, as comparing which sequence exhibits more motion generalizes across morphologies.

In inference (Fig.~\ref{fig:overview}, top), the second sequence is fixed to a known rest reference $X_{\text{rest}}$ (where the fingers are idle) and $X_t$ slides through the recording, yielding for each finger $f$ a continuous signal
$\hat{p}^{(f)}_t = P\bigl(\delta^{(f)}(X_t) > \delta^{(f)}(X_{\text{rest}}) \mid X_t, X_{\text{rest}}\bigr)$.
While this signal captures relative flexion magnitude, it does not directly indicate whether a finger is actively flexing, and a second stage is needed to turn the learned representations into real-time flexion decisions.

\begin{figure*}[t]
\centering
\includegraphics[width=0.975\textwidth]{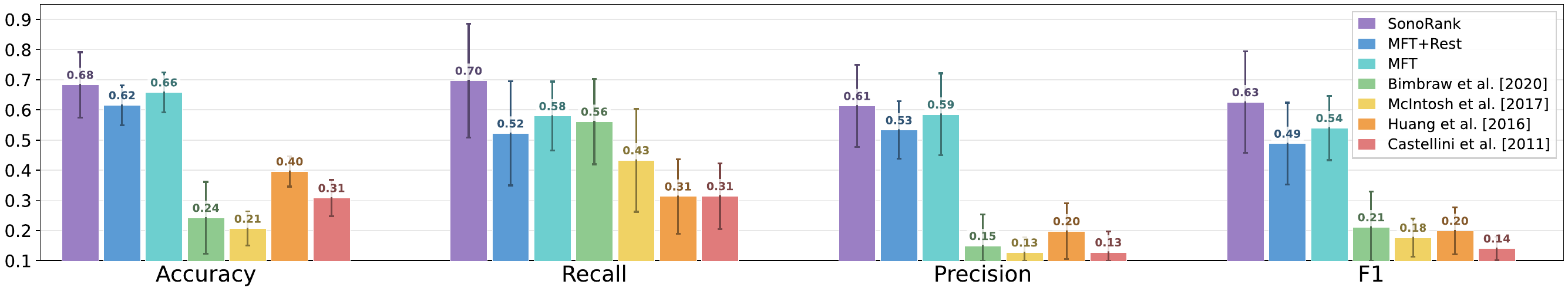}
\vspace{-0.2cm}
\caption{Accuracy, recall, precision, and F1 score for seven methods under $12$-fold leave-one-subject-out evaluation (error bars: $\pm 1$ std across folds).
The first three methods are ours: \SonoRank{} (full pipeline), \MFTRest{} (no ranking pretraining), and \MFT{} (no ranking, no rest reference).
The remaining four are external baselines reimplemented on our dataset~\cite{DBLP:conf/hsi/HuangL16,DBLP:conf/iros/CastelliniP11,DBLP:conf/chi/McIntoshMFP17,DBLP:journals/corr/abs-2211-15871}.
\SonoRank{} leads consistently across all four metrics, while external baselines achieve F1 scores of $0.21$ or lower, showing that single-frame classifiers and hand-crafted features fail to generalize across subjects.}
\label{fig:baselines}
\end{figure*}

\subsection{Flexion Classification}
\label{sec:flex_classification}

After ranking pretraining, we discard the five comparison heads and replace them with a classification head (Fig.~\ref{fig:overview}, bottom right): a multi-layer perceptron~(MLP) that maps sequence-level embeddings to five binary flex predictions.
Given a query sequence $X$ and a rest reference $X_{\text{rest}}$, both are encoded through the shared backbone (frame encoder, temporal transformer, and mean pooling) to produce sequence-level embeddings $\mathbf{e}$ and $\mathbf{e}_{\text{rest}}$.
The concatenated vector $[\mathbf{e},\mathbf{e}_{\text{rest}}]$ is passed through the MLP, which outputs a flex prediction $\hat{y}^{(f)} \in [0,1]$ for each finger~$f$.
Because the five predictions are produced independently rather than through a single mutually exclusive choice, the model can in principle represent several fingers flexing simultaneously.
The entire model, including the pretrained backbone, is fine-tuned end-to-end with $\ell_{\BCE}$ against ground-truth flex labels derived from the joint kinematics, where $y^{(f)} \in \{0, 1\}$ is the majority-vote flex label for finger~$f$ over the frames in the query window.
At inference, the window slides through the recording one frame at a time, producing a per-frame flexion decision for each finger.
This two-stage design separates representation learning from classification. Ranking teaches the backbone motion-sensitive features that generalize across subjects, and fine-tuning adapts them for flex detection.

\section{EXPERIMENTAL SETUP}
\label{sec:experiments}

\label{sec:dataset}

Our setup includes a Clarius L15HD (wireless B-mode ultrasound device) and a Vicon motion-capture system.
The ultrasound was configured for the musculoskeletal (MSK) mode at \SI{20}{\mega\hertz}, generating $480 \times 480$ images at 19 to 21 frames per second.
The probe was attached to the forearm in the ``transverse'' orientation using a wearable strap~\cite{DBLP:conf/chi/McIntoshMFP17}.
The Vicon system tracked reflective markers on the hand, from which joint angles across five fingers were extracted via a forward kinematic model~\cite{cerveri2007finger} (Appendix~\ref{sec:label_processing}).
Ultrasound images were downsampled to $224{\times}224$ and subsampled by a factor of two, yielding an effective rate of approximately \SI{10}{\hertz}.
We collected data from twelve right-handed subjects (seven men and five women, average age~23), all without neurological disorders or hand deformities.
Each subject completed three enrollment sessions of isolated finger flexions, with three to five recordings of approximately ninety seconds each and rest periods in between.
The final dataset comprises $51$ recordings totaling approximately $84$,$000$ ultrasound frames with synchronized kinematics.
The study was approved by the institution's Ethics Committee.


In all evaluations, we use 12-fold leave-one-subject-out cross-validation.
In each fold, one subject serves as the test set, the preceding subject as the validation set, and the remaining ten subjects form the training set.
The primary metric is per-finger area under the ROC curve (AUC), computed on active pairs ($m^{(f)}{=}1$).
We report mean AUC across the five fingers as the aggregate measure.
Training and optimization details are provided in Appendix~\ref{sec:model_implementation}.

\section{RESULTS}
\label{sec:results}

We first analyze \SonoRank{} through a series of ablations (Sec.~\ref{sec:ablation}): a comparison against direct classification baselines, an evaluation of different window lengths, and per-finger and per-subject breakdowns.
We then evaluate real-time inference throughput (Sec.~\ref{sec:realtime}).

\subsection{Ablation Study}
\label{sec:ablation}

\textbf{How important is ranking for generalizing to unseen subjects?}
We isolate the contribution of the ranking-pretrained backbone and the rest reference by comparing \SonoRank{} against two baselines (Fig.~\ref{fig:baselines}).
All three share the same architecture and initialize the image encoder from a model pretrained on image reconstruction~\cite{DBLP:conf/miccai/RonnebergerFB15}.
\MFT{} (Multi-Frame Transformer) trains the classifier directly on the pretrained encoder from the live sequence alone, and \MFTRest{} additionally receives the rest reference. Both skip the pairwise ranking stage that distinguishes \SonoRank{}.

\begin{figure*}[t]
\centering
\includegraphics[width=\textwidth]{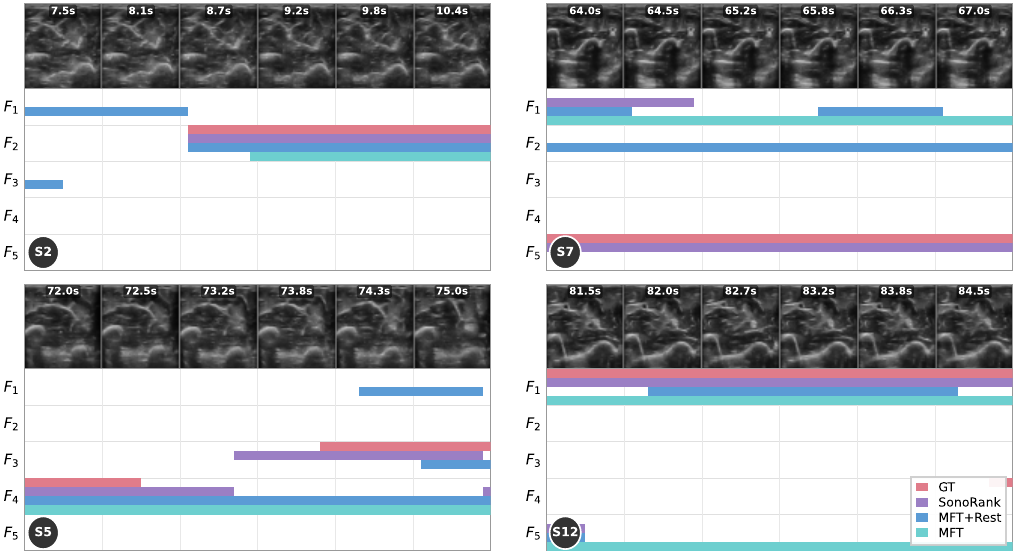}
\vspace{-0.5cm}
\caption{Representative examples of finger flexion predictions on four test subjects.
Each panel shows six ultrasound frames spanning a \SI{3}{\second} flexion interval (top) and the corresponding ground-truth and predicted flex labels for all five fingers (bottom).
(Top-left)~S2: clean $F_2$ (index) detection. \SonoRank{} matches the ground truth perfectly while baselines produce shorter or misaligned events.
(Top-right)~S7: $F_5$ (pinky) flexion that \SonoRank{} detects but both \MFTRest{} and \MFT{} miss entirely.
(Bottom-left)~S5: coupled $F_3$/$F_4$ (middle/ring) flexion. \SonoRank{} captures the dominant fingers but adds a spurious $F_4$ event.
(Bottom-right)~S12: $F_1$ (thumb) flexion, the hardest finger. All methods detect the event, yet \SonoRank{} aligns most closely with the ground truth.}
\vspace{-0.1cm}
\label{fig:qualitative}
\end{figure*}

\SonoRank{} achieves F1 of $0.63$, outperforming both \MFT{} ($0.54$) and \MFTRest{} ($0.49$), demonstrating the value of ranking pretraining.
A Wilcoxon signed-rank test yields $p{=}0.065$ with a 95\% CI of the mean difference of $[+0.001, +0.182]$.
That \MFTRest{} does not benefit from the rest reference without ranking pretraining indicates that the two components are synergistic. The backbone must learn \emph{how} to compare against rest for the reference to be useful.
Fig.~\ref{fig:qualitative} illustrates this on four test subjects: \SonoRank{} produces sharper, better-aligned detections, while baselines miss events or generate fragmented predictions.
Inter-subject variability remains high (F1 from $0.46$ to $0.83$, $\sigma{=}0.11$), underscoring the difficulty of cross-subject generalization.

To put the absolute performance of \SonoRank{} into context, we compare against four sonomyography methods reimplemented on our dataset under the same leave-one-subject-out protocol (Fig.~\ref{fig:baselines}).
Castellini and Passig~\cite{DBLP:conf/iros/CastelliniP11} extract features from circular image patches and predict finger positions with Ridge regression.
Huang and Liu~\cite{DBLP:conf/hsi/HuangL16} sample pixel-intensity columns and classify finger states with a two-layer MLP.
McIntosh et al.~\cite{DBLP:conf/chi/McIntoshMFP17} (EchoFlex) compute dense optical flow between consecutive frames and pool the resulting field for MLP classification.
Bimbraw et al.~\cite{DBLP:journals/corr/abs-2211-15871} classify individual frames using a modified VGG16.
All four achieve F1 of $0.21$ or lower on unseen subjects.
We also evaluated the CBMF model~\cite{DBLP:conf/icra/ZadokSWB23,zadok2025inferring}, but its training collapsed to constant outputs and was excluded.
Even \MFT{}, our weakest variant, outperforms all external baselines by a wide margin (accuracy $0.66$ vs.\ $0.40$), suggesting that multi-frame temporal modeling is a primary driver of cross-subject generalization.
The remaining ablations evaluate the ranking backbone directly using per-finger AUC, as the ranking stage determines the quality of the representations that the classifier builds upon.

\begin{figure*}[t]
\centering
\begin{subfigure}[b]{0.256\textwidth}
    \centering
    \includegraphics[width=\textwidth]{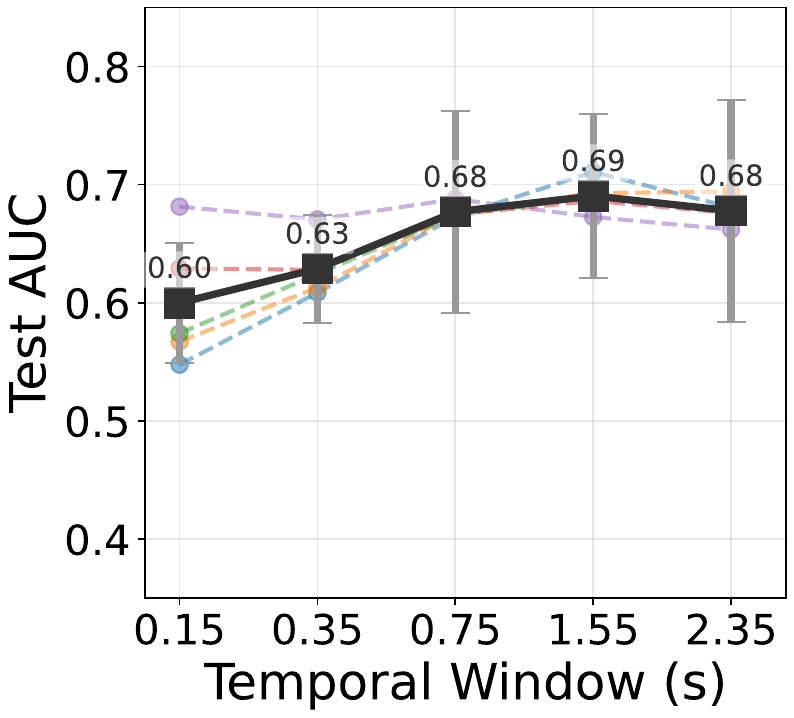}
    \vspace{-0.5cm}
    \caption{}
    \label{fig:seqlen_sweep}
\end{subfigure}%
\begin{subfigure}[b]{0.208\textwidth}
    \centering
    \includegraphics[width=\textwidth]{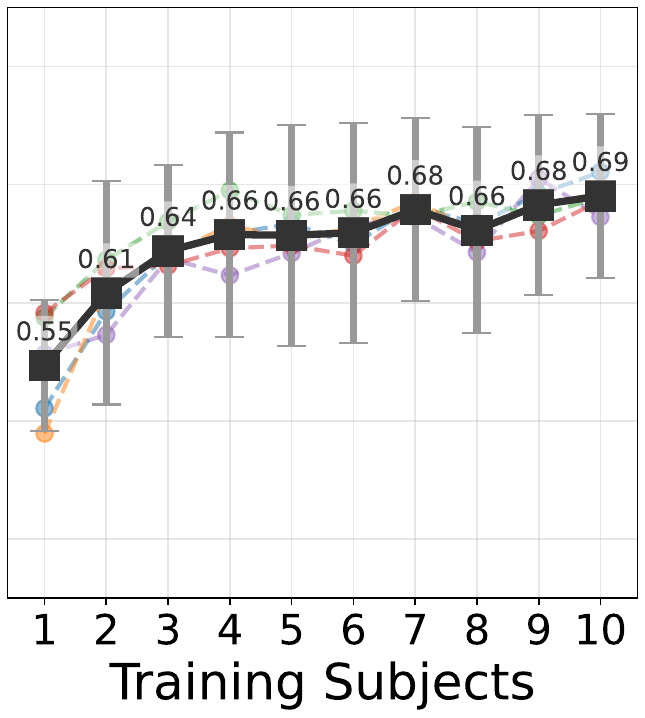}
    \vspace{-0.5cm}
    \caption{}
    \label{fig:subject_count}
\end{subfigure}%
\begin{subfigure}[b]{0.475\textwidth}
    \centering
    \includegraphics[width=\textwidth,trim=0 3px 0 0,clip]{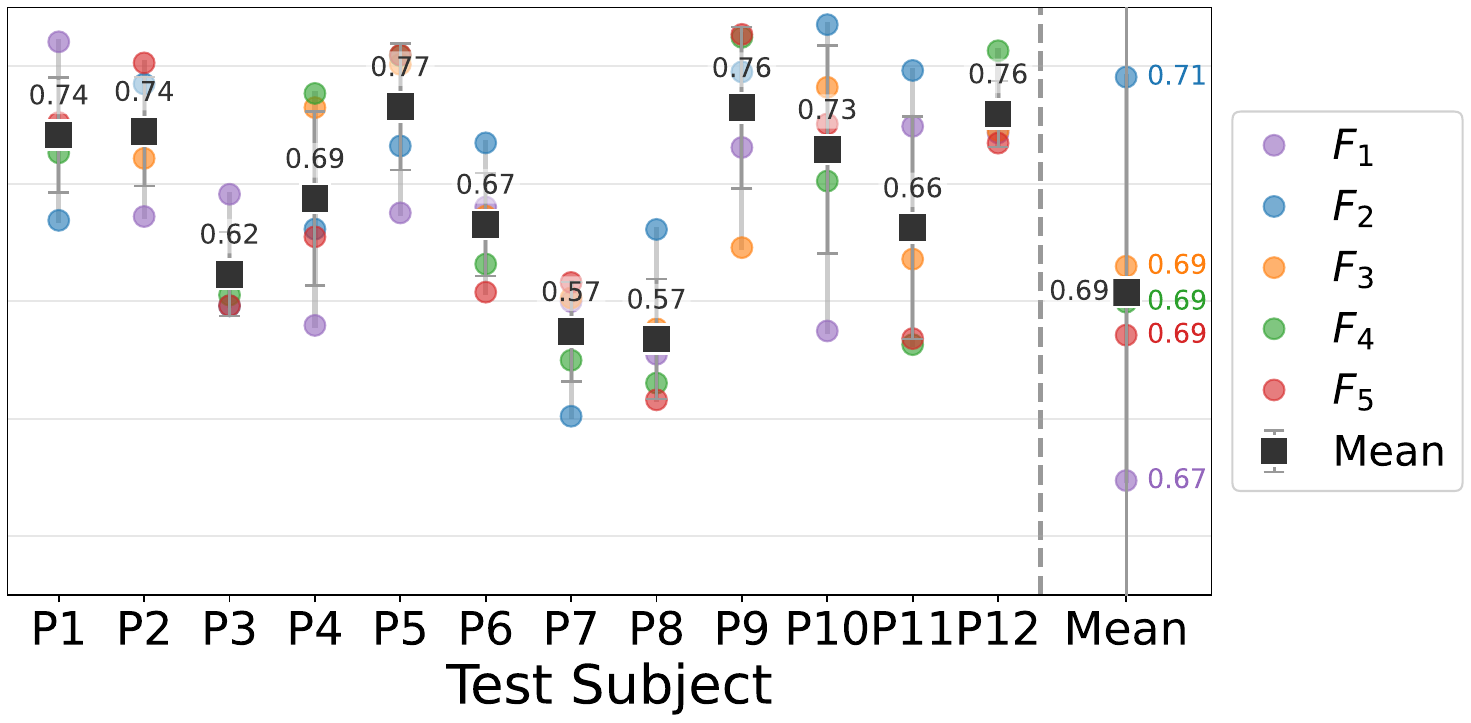}
    \vspace{-0.5cm}
    \caption{}
    \label{fig:subject_analysis}
\end{subfigure}
\vspace{-0.2cm}
\caption{
(a)~Temporal window sweep: mean per-finger ranking AUC as a function of window duration $T$ (from \SI{0.15}{\second} to \SI{2.35}{\second}), with dashed lines for individual fingers and black squares for mean values ($\pm 1$ std).
(b)~Training set size: ranking AUC as a function of the number of training subjects, showing that adding more than five subjects yields diminishing returns.
(c)~Per-subject ranking performance: each column shows per-finger AUCs (colored dots: $F_1$\,-\,$F_5$) and mean (black square). $F_2$ (index) leads at $0.71$, $F_1$ (thumb) trails at $0.67$, overall mean $0.69$.
All panels report mean ranking AUC averaged over $12$ leave-one-subject-out~folds.
}
\vspace{-0.2cm}
\label{fig:ablations}
\end{figure*}

\textbf{How much temporal context is needed to capture finger flexion?}
To evaluate the effect of context length on ranking quality, we measure per-finger AUC across different window durations $T$ (Fig.~\ref{fig:ablations}a).
Performance climbs steeply from $T{=}\SI{0.15}{\second}$ ($2$ frames) through $T{=}\SI{0.75}{\second}$ ($8$ frames) and peaks at $T{=}\SI{1.55}{\second}$ ($16$ frames), roughly spanning one flexion-extension cycle.
Extending to $T{=}\SI{2.35}{\second}$ ($24$ frames) yields no further gain, and longer windows were infeasible due to GPU memory constraints.
The per-finger breakdown reveals an interesting reversal: at $T{=}\SI{0.15}{\second}$ $F_1$ (thumb) is the easiest finger to discriminate ($0.68$) while $F_2$ (index) is the hardest ($0.55$), yet at $T{=}\SI{1.55}{\second}$ this ranking inverts, with $F_2$ reaching $0.71$ and $F_1$ dropping to $0.67$.
We attribute this to the quicker flexion motion of $F_1$, which is easier to capture in a shorter window, whereas the other fingers produce longer motions that benefit from longer context.

\textbf{How does prediction difficulty vary across subjects and fingers?}
Fig.~\ref{fig:ablations}c reveals large inter-subject variability: the top four subjects (S5, S9, S12, S2) average $0.76$, while the bottom four (S11, S3, S7, S8) average $0.61$.
We attribute this spread to differences in forearm morphology and subject performance during data collection.
Averaged across subjects, $F_2$ (index) is the easiest finger to predict ($0.71$), followed by $F_4$ (ring) and $F_5$ (pinky) whose shared FDP tendons cause overlapping ultrasound signatures, while $F_1$ (thumb) ranks lowest ($0.67$).
We attribute this gap to $F_1$ anatomy (Sec.~\ref{sec:method}): its extrinsic flexor (FPL) is one of the smallest muscles in the image, and the primary thumb flexors lie in the palm, outside the ultrasound field of view.
Nevertheless, finger difficulty varies across subjects: $F_1$ is the top-performing finger for S1 ($0.82$) yet the weakest for S4 ($0.58$), indicating that individual anatomy and flexing behavior affect which fingers are easiest to decode.

\textbf{How many subjects are needed for the ranking model to generalize?}
Fig.~\ref{fig:ablations}b shows ranking AUC as a function of the number of training subjects.
Performance rises steeply from one subject ($0.55$) to three ($0.64$).
However, adding subjects beyond five yields only marginal gains, with the entire set of ten subjects reaching $0.69$.
This suggests that approximately five subjects already capture sufficient forearm morphology diversity for the ranking model to generalize.
This is an encouraging finding for practical deployment where data collection is costly.

\subsection{Real-Time System Performance}
\label{sec:realtime}

\begin{figure}[t]
\centering
\includegraphics[width=0.45\textwidth]{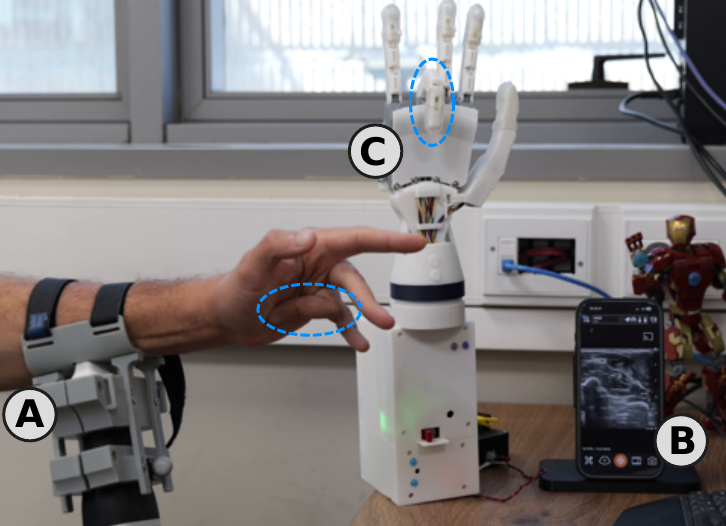}
\vspace{-0.05cm}
\caption{Real-time demonstration of the \SonoRank{} system.
(A)~Clarius ultrasound probe mounted on the forearm with a 3D-printed holder.
(B)~Smartphone displaying the live ultrasound feed (for visualization only). Frames are streamed directly from the probe to the PC.
(C)~Robotic hand~\cite{zadok2026digiarm} mirroring the detected finger flexion in real time.}
\vspace{-0.4cm}
\label{fig:realtime_demo}
\end{figure}

To demonstrate that \SonoRank{} can operate in real time, we built an end-to-end system (Fig.~\ref{fig:realtime_demo}) connecting a Clarius L15HD ultrasound probe to a robotic hand~\cite{zadok2026digiarm} via a PC running inference.
The demonstration was performed with a new user who did not participate in the offline data collection and had no prior experience operating the system.
The startup procedure requires only that the user hold their hand relaxed for two to three seconds.
During this brief period, the system captures a single rest frame, which is duplicated $T{=}16$ times to form the rest reference $X_{\text{rest}}$ that remains fixed for the entire session.
Simultaneously, the sliding buffer of $31$ raw frames fills as the probe streams at \SI{20}{\hertz}, and the model observes the baseline per-finger probabilities during rest to automatically set activation thresholds.
Once the buffer is full, each new frame triggers a forward pass that produces five per-finger flex probabilities.
The system selects the single most confident finger above its threshold and sends a command to the robotic hand, filtering brief fluctuations to ensure stable control.

As for model latency, we benchmarked the inference pipeline on a consumer-grade NVIDIA RTX~3080 using pre-recorded H5 sessions.
The full model has \num{10.7} million parameters, and a single forward pass takes \SI{17.5}{\milli\second} on average, corresponding to \SI{57}{\hertz}, nearly three times the \SI{20}{\hertz} ultrasound frame rate.
Because the classifier is trained with majority-vote labels, a flexion is typically detected once roughly half the window contains flexion frames, yielding an estimated detection delay of ${\sim}\SI{0.78}{\second}$.
This delay can be reduced to ${\sim}\SI{0.39}{\second}$ by halving the temporal window, which retains ranking performance (Fig.~\ref{fig:ablations}a).
In practice, the observed algorithmic delay was ${\sim}\SI{0.5}{\second}$, confirming real-time operation.
Note that the actuation delay, which depends on the robotic platform, is added on top.
Since our method requires only an update rate of \SI{10}{\hertz}, the computational prediction budget is \SI{100}{\milli\second}, leaving plenty of room for deployment on embedded platforms.
Techniques such as half-precision inference and model pruning can reduce latency further without modifying the architecture.

\section{DISCUSSION AND FUTURE WORK}
\label{sec:discussion}

We presented \SonoRank{}, a two-stage approach for detecting per-finger flexion from forearm ultrasound without subject-specific calibration.
The results show that temporal modeling and pairwise ranking are complementary drivers of cross-subject generalization: temporal context alone (\MFT{}) already outperforms external baselines, while the ranking-pretrained backbone combined with a rest reference (\SonoRank{}) lifts macro F1 to $0.63$, roughly three times that of the best external baseline.
The model runs at \SI{57}{\hertz} on a consumer-grade GPU with sub-second detection delay (Sec.~\ref{sec:realtime}).
These results notwithstanding, several limitations remain.
The study is limited to twelve healthy, right-handed subjects performing isolated single-finger flexions in a controlled setting.
Inter-subject variability is high (F1 scores from $0.46$ to $0.83$, $\sigma{=}0.11$), and $F_1$ (thumb) is consistently the hardest finger to detect, likely because its primary flexors reside in the palm and are invisible to forearm ultrasound (Fig.~\ref{fig:ablations}c).

This variability indicates that current accuracy is not yet sufficient for production.
Defining a minimal performance threshold, below which the system requests a calibration session, could close this gap while preserving calibration-free operation for most users.
Beyond prosthetics, calibration-free finger detection can enable engagement with digital devices~\cite{chen2025robotic}, musical instruments~\cite{DBLP:conf/biorob/CastelliniHSGN14}, gaming~\cite{4696889}, and virtual reality~\cite{sgambato2025vr} without per-session setup.
Most importantly, because the ranking formulation learns relative motion comparisons rather than absolute patterns, it is well-suited for generalization to individuals with upper-limb amputations, whose residual muscle structure differs from intact forearms. Validating this is the next step toward intuitive and calibration-free prosthetic finger control.

\section*{ACKNOWLEDGMENT}

We thank Haifa3D for their generous support and shared vision of accessible assistive technology.
This project has received funding from the Israeli Ministry of Science \& Technology (grant No.~8774), the David Himelberg Foundation, and the Wynn Family Foundation.




\appendix
\section{}

\subsection{Label Processing}
\label{sec:label_processing}

\begin{figure}[t]
\centering
\includegraphics[trim=0 20 0 25, clip, width=0.275\textwidth]{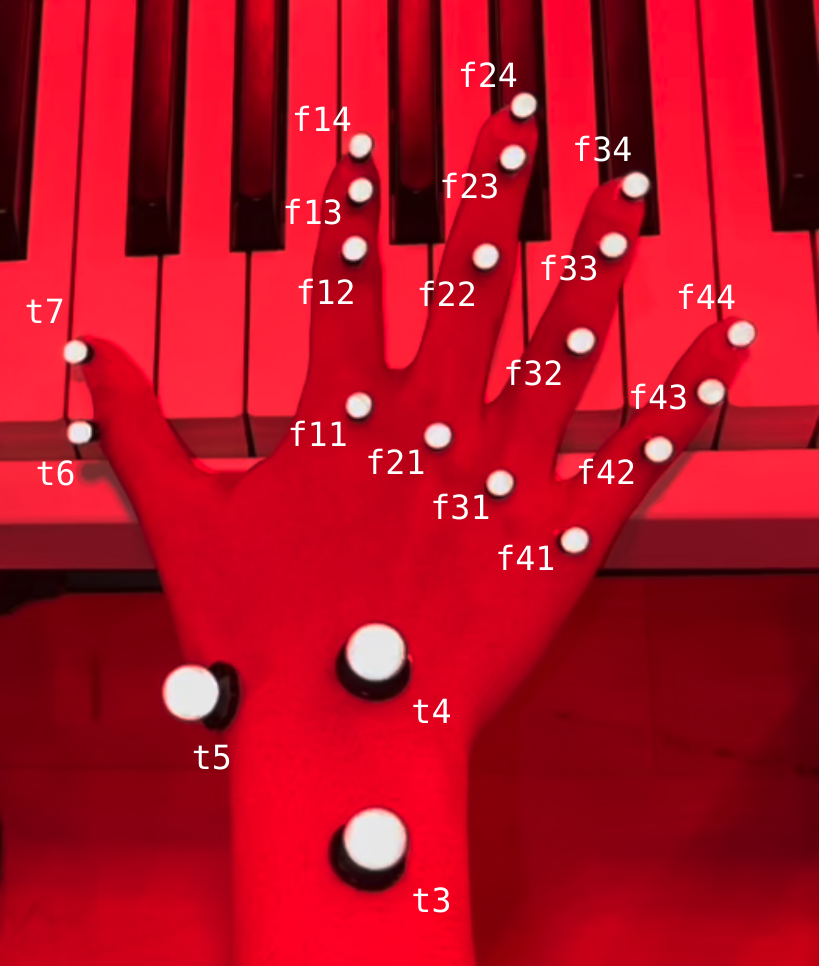}
\caption{Position of each reflective marker on the subject's hand, as tracked by the Vicon motion-capture system.}
\vspace{-0.25cm}
\label{fig:markers_sample}
\end{figure}

The joint angles used to define flexion labels are computed from reflective hand markers tracked by a Vicon motion-capture system (Fig.~\ref{fig:markers_sample}).
Each marker provides a 3D position, and we denote the vector from marker $a$ to marker $b$ as $\vec{v}_{a \rightarrow b} = \vec{b} - \vec{a}$.
The angle at a joint is computed from two vectors sharing a common vertex using the two-argument arctangent:
\begin{equation}
  \angle(\vec{u}, \vec{v}) = \arctantwo\!\bigl(\norm{\vec{u} \times \vec{v}}_2,\; \vec{u} \cdot \vec{v}\bigr).
  \label{eq:angle}
\end{equation}

Each of the four fingers ($F_2$ to $F_5$) has three anatomical joints: metacarpophalangeal (MP), proximal interphalangeal (PIP), and distal interphalangeal (DIP).
In this work, we compute only the MP and PIP angles, as these two joints capture the dominant flexion motion.
Using the little finger ($F_5$) as an example, with markers $f_{41}$ (MP), $f_{42}$ (PIP), and $f_{43}$ (DIP tip), and $t_5$ on the metacarpal:
\begin{equation}
\begin{aligned}
  \theta^{(F_5)}_{\text{MP}}  &= \angle\!\bigl(\vec{v}_{f_{41} \rightarrow t_{5}},\; \vec{v}_{f_{41} \rightarrow f_{42}}\bigr), \\
  \theta^{(F_5)}_{\text{PIP}} &= \angle\!\bigl(\vec{v}_{f_{42} \rightarrow f_{41}},\; \vec{v}_{f_{42} \rightarrow f_{43}}\bigr).
\end{aligned}
\label{eq:finger_angles}
\end{equation}
The thumb ($F_1$) is handled differently because its pressing motion is not well described by the same planar angles.
The interphalangeal joint $\theta^{(F_1)}_{\text{PIP}}$ is the angle at marker $t_6$ between $\vec{v}_{t_6 \rightarrow t_7}$ and $\vec{v}_{t_6 \rightarrow t_5}$, while the MP joint $\theta^{(F_1)}_{\text{MP}}$ is the angle between $\vec{v}_{t_5 \rightarrow t_6}$ and the normal $\hat{\vec{n}}_p$ to the palm plane, estimated from markers $f_{11}$, $f_{41}$, and $t_4$.

The total flexion angle for finger $f$ is $\theta^{(f)}_t = \theta^{(f)}_{\text{MP},t} + \theta^{(f)}_{\text{PIP},t}$ (Sec.~\ref{sec:formulation}).
A frame is labeled as flexing if $\theta^{(f)}_t$ exceeds a threshold derived from each finger's range of motion during the session.
For the ranking stage, the motion magnitude $\delta^{(f)}(X)$ (Eq.~\ref{eq:delta}) provides the pairwise labels.
For the classification stage, the per-sequence flex label is the majority vote over frames in the window.


\subsection{Model Implementation}
\label{sec:model_implementation}


All three models evaluated in Sec.~\ref{sec:ablation} (\SonoRank{}, \MFTRest{}, and \MFT{}) share the same backbone architecture.
The backbone consists of a frame encoder followed by a temporal transformer.
The frame encoder is a five-stage convolutional network with residual connections.
The first stage applies a $7 \times 7$ convolution with stride~2 to the $224 \times 224$ input, producing 64 feature maps with a residual block, stages~2 to~4 each downsample with stride-2 convolutions and double the channel count (128, 256, 512) with residual blocks, and stage~5 applies a final stride-2 convolution (512 channels).
Global average pooling and a linear layer project the resulting feature map to a 64-dimensional embedding vector.
Each frame in the input sequence is encoded independently using the same shared weights.
The sequence of frame embeddings is then summed with a learnable positional encoding and passed through a Transformer encoder with 2 layers, 4 attention heads, a feed-forward dimension of 256, and pre-layer normalization.
Mean pooling over the output sequence yields a 64-dimensional embedding of the input window's temporal dynamics.

In \SonoRank{}, the backbone is first trained with five parallel ranking heads (Sec.~\ref{sec:ranking}).
Each ranking head is a four-layer MLP (128$\to$256$\to$128$\to$64$\to$1) with ReLU and dropout, mapping the concatenated embeddings of two sequences to a scalar logit $P(\text{flex}_A > \text{flex}_B)$.
After ranking pretraining, the five heads are replaced by a classification MLP (128$\to$128$\to$64$\to$5) mapping the concatenated query and rest embeddings to five independent flex probabilities (Sec.~\ref{sec:flex_classification}).
Only the classification head is initialized from scratch, and the full model is fine-tuned end-to-end.

\MFTRest{} uses the same classification MLP and rest-reference input as \SonoRank{}, but skips the ranking stage entirely: the backbone is initialized from an image-reconstruction pretraining, and the classifier is trained directly.
\MFT{} further removes the rest reference, so its classification MLP receives only the 64-dimensional query embedding (64$\to$128$\to$64$\to$5) and must predict flexion from the live sequence alone.

The ranking stage is trained for 10 epochs with a batch size of~8 and $5 \times 10^4$ virtual pair samples per epoch.
We use the Adam optimizer with discriminative learning rates: $10^{-5}$ for early encoder layers, $5 \times 10^{-5}$ for late encoder layers, $10^{-4}$ for the transformer, and $3 \times 10^{-4}$ for the comparison heads, with cosine annealing.
The uncertainty penalty coefficient (Eq.~\ref{eq:loss}) is set to $\lambda{=}0.5$, selected via grid search over $\{0, 0.1, 0.25, 0.5, 1, 2\}$. The ranking performance was stable for $\lambda \leq 0.5$ and degraded for larger values.
The classification stage then fine-tunes the full model end-to-end for $10$ additional epochs.
Dropout is set to~$0.3$ throughout.
Geometric and intensity augmentations are applied consistently across all frames in each sequence to preserve temporal coherence.
The full implementation, including training configurations, is available at \textcolor{blue}{\href{https://github.com/deanzadok/sonorank}{github.com/deanzadok/sonorank}}.


\bibliographystyle{IEEEtran}
\bibliography{main}

\end{document}